\pdfoutput=1
\RequirePackage{fix-cm}
\documentclass[twocolumn]{svjour3}
\smartqed

\usepackage{graphicx}
\usepackage{url}
\usepackage{xcolor}
\usepackage{hyperref}
\hypersetup{
     colorlinks=true,
     linkcolor=orange,
     filecolor=orange,
     citecolor=orange,      
     urlcolor=orange,
     }
\usepackage{color, colortbl}
\usepackage{amsfonts}
\usepackage{amsmath}
\usepackage{appendix}
\usepackage{algorithm, setspace}
\usepackage{algpseudocode}
\usepackage{csquotes}
\usepackage{amsfonts}
\usepackage{booktabs}
\usepackage{siunitx}
\usepackage{breqn}
\usepackage[normalem]{ulem}
\usepackage{tikz}
\usetikzlibrary{tikzmark}

\newcommand{\p}[1]{\medskip \noindent \textbf{{#1}.}}
\newcommand{\eq}[1]{Equation~(\ref{eq:#1})}
\newcommand{\fig}[1]{Figure~\ref{fig:#1}}

\DeclareMathOperator*{\argmin}{arg\,min}

\journalname{Autonomous Robots}
\DeclareUnicodeCharacter{2212}{-}

\begin{document}

\title{L2D2: Robot Learning from 2D Drawings}
\author{Shaunak A. Mehta \and Heramb Nemlekar \and Hari Sumant \and Dylan P. Losey}

\institute{S.A. Mehta \at
              Mechanical Engineering, Virginia Tech \\
              \email{mehtashaunak@vt.edu}           
           \and
           H. Nemlekar \at
              Mechanical Engineering, Virginia Tech \\
              \email{hnemlekar@vt.edu}           
            \and
           H. Sumant \at
              Electrical and Computer Engineering, Virginia Tech \\
              \email{harisumant@vt.edu}           
           \and
           D.P. Losey \at
            Mechanical Engineering, Virginia Tech \\
            \email{losey@vt.edu}
}

\maketitle

\begin{abstract}

Robots should learn new tasks from humans.
But how do humans convey what they want the robot to do?
Existing methods largely rely on 
humans physically guiding 
the robot arm throughout their intended task.
Unfortunately --- as we scale up the amount of data --- physical guidance becomes prohibitively burdensome.
Not only do humans need to operate robot hardware
but also modify the environment (e.g., moving and resetting objects) to provide multiple task examples.
In this work we propose L2D2, a sketching interface and imitation learning algorithm where humans can provide demonstrations by \textit{drawing} the task.
L2D2 starts with a single image of the robot arm and its workspace.
Using a tablet, users draw and label trajectories on this image to illustrate how the robot should act.
To collect new and diverse demonstrations, we no longer need the human to physically reset the workspace; instead, L2D2 leverages vision-language segmentation to autonomously vary object locations and generate synthetic images for the human to draw upon.
We recognize that drawing trajectories is not as information-rich as physically demonstrating the task.
Drawings are 2-dimensional and do not capture how the robot's actions affect its environment.
To address these fundamental challenges the next stage of L2D2 grounds the human's static, 2D drawings in our dynamic, 3D world by leveraging a small set of physical demonstrations.
Our experiments and user study suggest that L2D2 enables humans to provide more demonstrations with less time and effort than traditional approaches, and users prefer drawings over physical manipulation.
When compared to other drawing-based approaches, we find that L2D2 learns more performant robot policies, requires a smaller dataset, and can generalize to longer-horizon tasks.
See our project website: \url{https://collab.me.vt.edu/L2D2/}

\end{abstract}


\section{Introduction}



\begin{figure}[t]
    \centering
    \includegraphics[width=1\linewidth]{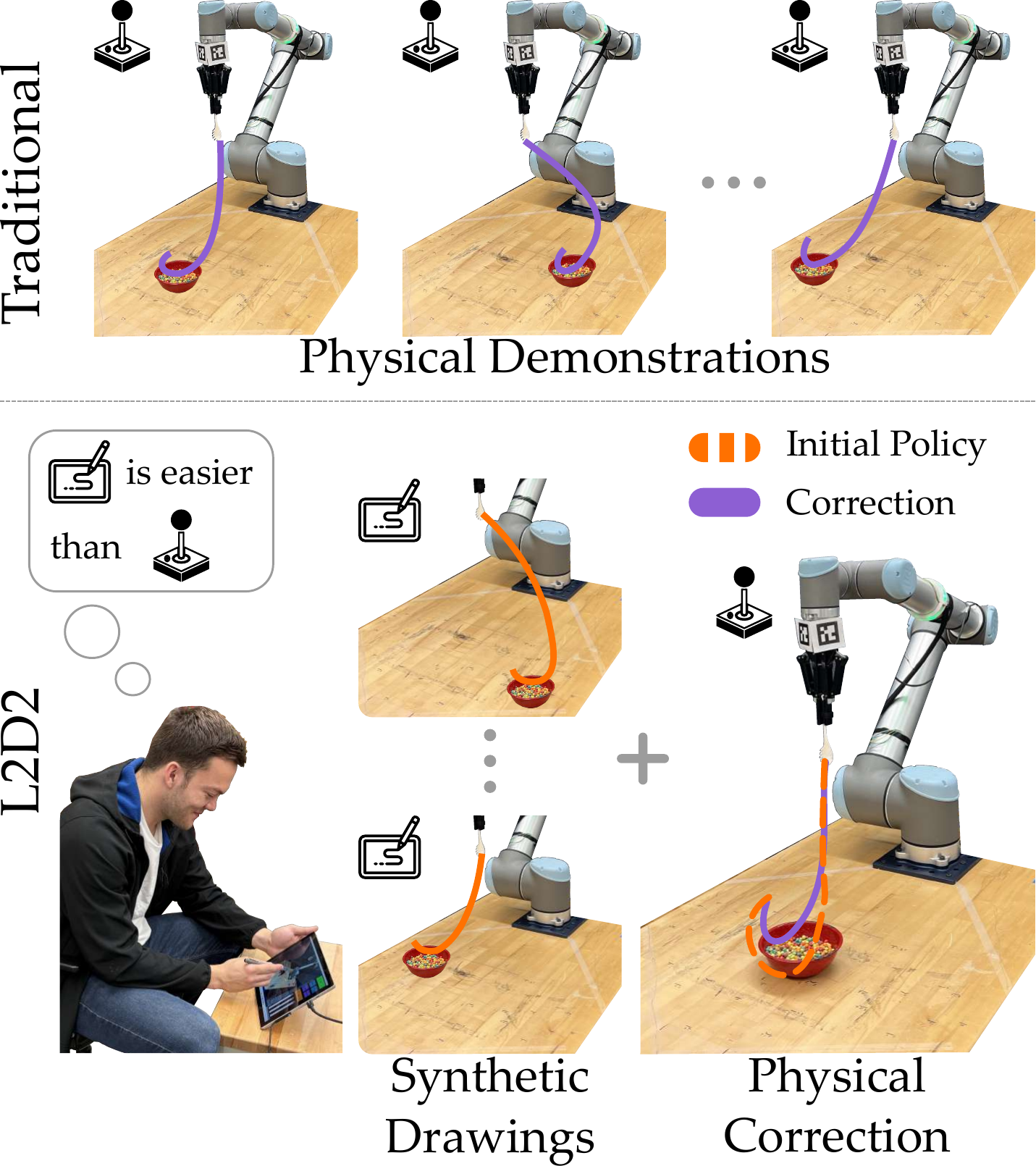}
    \caption{Human demonstrating a scooping task to the robot using different teaching paradigms. When using traditional methods to teach the robot, 
    the user needs to manually reset the environment by changing the bowl position and provide demonstrations by physically guiding the robot through the task. We propose L2D2, an approach that synthetically generates diverse environment settings and enables the human to demonstrate the task by drawing a trajectory on the artificial images of the environment. If the robot makes a mistake when executing the learned task, the user provides a few physical corrections to fine-tune the robot's learned behavior.
    Our proposed approach reduces costly physical interactions and enables humans to teach robots efficiently.}
    \label{fig:front}
\end{figure}

Robots can learn new tasks by imitating human examples. In general, as humans provide more examples of a task, and the more diverse those examples are, the robot will learn to perform that task more effectively.  
Consider the task depicted in Figure \ref{fig:front}, where the robot must scoop cereal from a bowl. One common way for humans to teach this task is by physically demonstrating it, i.e., kinesthetically backdriving the robot's joints through the process of reaching the bowl and rotating the spoon. 
The robot can simply copy this motion if the bowl always stays in the same place. But for the robot to learn how to scoop cereal when the bowl is moved, the human needs to demonstrate the task for various bowl positions. 
This process of teaching the robot can not only be challenging --- the human needs to carefully orchestrate the motion of the robot --- but is also time-consuming. 
Each time the human wants to show another example, they will have to reset the robot, place the bowl in a new spot, and demonstrate the task again. 
So how can we make it easy for humans to provide these diverse examples?

Recent works have explored methods that simplify the teaching process by enabling humans to conveniently provide informative demonstrations. For example, humans can control the robot via remote teleoperation, making the process as intuitive as performing the task with their own hands ~\cite{sian2004whole,jonnavittula2024sari,liu2024libero}. Alternatively, humans can take a video where they perform the task themselves and provide language descriptions that explain their actions \cite{jonnavittula2025view,jain2024vid2robot,alakuijala2023learning,bahl2022human}. 
These works follow a general trend: 1) making it easier for humans to perform the task and 2) accurately capturing the task complexities through real-world demonstrations. However, for these existing approaches to work, the human still needs to perform the task either vicariously or with their own body while physically changing the environment for every example they demonstrate.
Because each iteration of providing an example is laborious, this fundamentally limits how much these approaches can accelerate data collection.

We want to make it easier for humans to teach the robot. Building on the recent trends, we envision a system where the human can teach the robot by \textit{drawing} the desired task on an image of the environment. Humans can generate drawings rapidly, and --- because they are not actually performing the task --- the human teacher is not constrained by the physical speed of the learner or the burden of resetting the environment between demonstrations. At the same time, we recognize that drawings as a form of demonstration are themselves limited: when using drawings the human is potentially trying to convey a complex, high-dimensional task on a $2$D image of the environment. To bridge the gap between the low-cost, easy to provide drawings and information-rich demonstrations, we hypothesize that:
\begin{center}
    \textit{Robots can rapidly learn new tasks by\\ combining diverse canonical drawings with a few high-dimensional demonstrations.}
\end{center}

We propose \textbf{L2D2}: \textbf{L}earning from \textbf{2D D}rawings, a three-step approach for learning from human teachers. First, the robot takes an image of the environment, and the human annotates task-relevant objects or features that can vary between task iterations. Next, the human teacher iteratively draws the task on the images provided by the robot: at each iteration, we leverage vision and language models to artificially manipulate the positions of the objects in the image and adjust the features that the human highlighted. This approach quickly results in a large dataset of diverse demonstrations. For instance, in our user study, in the time it takes to provide $10$ physical demonstrations, users were able to provide ${\sim}20$ drawings. 

In the final step, the robot extracts high-dimensional demonstrations from the drawings, trains a control policy, and then attempts to perform the task in the real world by rolling out the learned policy. If the human teacher sees that the robot is making a mistake when performing the task, they can physically correct the robot to refine its behavior. 
These corrections help the robot ground the drawings in reality and fill any gaps in information.
Returning to our motivating example in Figure \ref{fig:front}, after observing the diverse drawings, the robot may learn to reach the cup but may not learn to precisely rotate the spoon. The human teacher can demonstrate this rotation at run-time to correct the robot's motion, thus reinforcing the learned policy.

Overall, we make the following contributions:


\p{Sketching Interface}
We create an interface that people can use to provide information-rich sketches to teach the robot a desired task. This interface presents a third-person image of the environment, and enables users to draw the robot's trajectory, annotate where the robot should rotate, open or close its gripper, and also highlight environment features that may vary at test time.


\p{Quantifying Information Loss in Sketches}
We analyze the process of decoding task sketches into real robot trajectories using Principal Component Analysis (PCA). We model the information loss as a function of the viewpoint from which we capture the images that users draw on.
Leveraging this model, we propose a strategy for optimally placing the camera to minimize information loss, and establish an upper bound on this loss when we use non-linear mappings between 3D robot trajectories and 2D sketches.


\p{Obtaining Information-Rich Drawings}
Given the importance of varied examples in imitation learning and the difficulty of collecting them physically, we leverage vision-language models to synthetically augment the data collection process. Specifically, we enable users to identify relevant features (e.g., target objects) through text prompts and autonomously vary those features with each drawing. 
Our approach results in a more diverse training set without the need for physically manipulating the environment. 
That is, instead of presenting the same image repeatedly, we change the locations of objects across images, enabling the user to show different examples of how to perform the task. 


\p{Incorporating Physical Feedback}
Recognizing that sketches themselves are not sufficient to learn dynamic, high-dimensional tasks due to information gaps in static, low-dimensional sketches, we utilize a few physical corrections to refine user drawings as well as the behaviors learned from them. This helps the robot ground its synthetic sketches in the physical environment and learn the dynamic task interactions more precisely.

Putting all parts together, our resulting approach, L2D2, integrates drawings and physical feedback in a way that amplifies their complementary strengths: using physical corrections to improve sketch accuracy, resulting in data that is both varied and reliable.


\p{Comparing with Baselines}
We conduct experiments to compare our approach to state-of-the-art approaches in learning from drawings and standard baselines that learn from physical demonstrations.
These experiments cover multiple real-world manipulation tasks performed both by expert users and by novice users as part of a user study.
Our results show that the data collected by L2D2 enhances the robot's task performance compared to alternative sketching methods. It also requires significantly less time and effort than traditional data collection approaches, which involve physically modifying the environment and demonstrating the task.

\section{Related Work}




Our work focuses on efficiently collecting diverse and accurate data for training robots to perform new tasks via imitation learning. In particular, we propose obtaining this information through sketches drawn on synthetically generated images to reduce human effort. In this section, we review prior techniques and interfaces for streamlining data collection in imitation learning. 

\p{Imitation Learning}
Imitation learning enables robots to learn new tasks by mimicking the behavior of an expert human. 
A widely used imitation learning method in robotics is behavior cloning (BC)~\cite{pomerleau1988alvinn,mehta2025stable} in which the robot is provided a dataset of observation-action pairs for some task. From this dataset, the robot learns a policy that takes the current task observation as input and predicts an action that leads to the successful completion of the task~\cite{osa2018algorithmic}.

A key factor in the effectiveness of this approach is the quality and variety of human data~\cite{belkhale2023data}. 
For example, to learn how to pick up a cup from various locations, the robot must be shown how an expert performs the task in at least a few different positions.
But obtaining this data can require significant human effort in setting up the environment and demonstrating the task~\cite{ravichandar2020recent}.
Previous research has attempted to address this problem from different angles: making it easy for humans to demonstrate the task~\cite{fang2024airexo}, synthetically augmenting the human's data~\cite{zhang2024diffusion}, and developing data-efficient learning algorithms~\cite{jang2022bc}. 
In this work, we utilize existing learning rules and investigate frameworks for obtaining the required training examples with novel interfaces and data augmentation techniques.

\p{Interfaces for Providing Demonstrations}
Humans can demonstrate their desired task by guiding the robot through the task~\cite{sian2004whole,liu2024libero,mehta2024unified,wu2024gello,iyer2024open} or by performing the task themselves~\cite{song2020grasping,jonnavittula2025view}.
Each approach has its own advantages; while performing the task directly is more natural for humans, controlling the robot results in data that is easier for the robot to learn from. Below, we discuss these approaches in more detail.

Traditional approaches for guiding the robot include kinesthetically moving the robot~\cite{argall2009survey} or teleoperating the robot using a joystick~\cite{sian2004whole,jonnavittula2024sari} or a space mouse~\cite{liu2024libero}. While these interfaces are easily accessible, they are not always intuitive to use when controlling high-dimensional robots. 
To make this process more intuitive, recent works have developed intelligent mappings for existing interfaces~\cite{mehta2022learning} as well as new tools like dexterous robot arm copies~\cite{fu2024mobile,wu2024gello,fang2024airexo} and virtual reality systems~\cite{zhang2018deep,iyer2024open,george2025openvr} for teleoperating the robot from an egocentric view.

Humans do not need to consider these interfaces when performing the task themselves. Robots can learn to perform diverse tasks by observing the human's behavior from a third-person camera~\cite{jonnavittula2025view,jain2024vid2robot,alakuijala2023learning,bahl2022human}.
However, since this data does not include observations and actions from the robot's viewpoint, it needs to be processed to extract relevant data.
For this reason, other approaches have developed reacher-grabber tools equipped with onboard cameras that humans can hold to perform tasks from the robot's perspective~\cite{young2021visual,song2020grasping,shafiullah2023bringing}. 

A limiting factor of both these teaching modalities is that they require humans to either directly or vicariously interact with the physical environment. 
Further, despite the method employed to obtain demonstrations, humans still need to manually vary the environment in order to introduce diversity in the training data.

\p{Data Augmentation}
To acquire sufficient data with only a few physical interventions, robots can leverage open-source videos of humans performing the task~\cite{bharadhwaj2024track2act} or large-scale cross-domain robot datasets~\cite{ebert2021bridge,brohan2022rt,o2024open}, and then transfer their knowledge to the desired task~\cite{dasari2021transformers,james2018task,duan2017one,jang2022bc}.
Recent advances have also utilized large language models to facilitate this transfer through contextual language prompts~\cite{lynch2020language,nguyen2019vision,stepputtis2020language}.
Alternatively, robots can collect a small amount of data from humans and augment it synthetically --- either by adding noise to human actions and then taking corrective actions to return to the demonstrated states~\cite{laskey2017dart,hoque2023interventional}, or by introducing semantic variations in the robot's observation~\cite{pari2022surprising,mandi2022cacti,yu2023scaling}, or both~\cite{zhang2024diffusion}.
For instance, robots can employ pre-trained vision-language models to segment out regions of interest in images of the environment, and in-paint those regions with different objects and backgrounds to make the policy robust to scene changes~\cite{mandi2022cacti}.

While these approaches help the robot generalize to unseen but familiar tasks and adapt to minor changes in behaviors and visual scenes, they do not generate new expert data. For example, the robot may learn to pick the cup from the same position across visually different environments, but it would not be able to pick the cup from a new location that requires a distinct motion.

\p{Learning from Sketches}
In this paper, we propose leveraging sketches to quickly generate diverse data for training robots.
Upcoming research in imitation learning has explored collecting demonstrations in the form of sketches of the desired behavior~\cite{sundaresan2024rt,gu2023rt,tanada2024sketch,zhi2024instructing,shah2012sketch,yu2025sketch}. In these approaches, users see an image of the environment and provide drawings that demonstrate the task. 
These drawings are then mapped to real robot trajectories~\cite{zhi2024instructing} or used to condition the robot's behavior~\cite{gu2023rt}.
For instance, \cite{yu2025sketch} collects 2D trajectories drawn on images taken from two viewpoints and generates the corresponding 3D trajectory using a pre-trained autoencoder framework, while \cite{gu2023rt} uses the drawing as input to the robot's policy and outputs the actions for rolling out the trajectory in the real world.

While sketches offer an intuitive way for humans to demonstrate the desired task without physically interacting with the environment, previous work has limited users to drawing on only real-world images, requiring them to reset the environment and collect new images before providing new sketches for the task.
Our work recognizes the potential of using sketches to rapidly generate large amounts of diverse training data and applies insights from data augmentation to synthetically create varied images of the task. We then obtain expert drawings on these synthetic images, capturing distinct task behaviors.
Recognizing the limitations of sketches in conveying precise and dexterous motions, we ground them with a small number of physical demonstrations to produce a dataset that is both varied and high quality. In our experiments, we show that by combining synthetic data generation and physical feedback, we can enable the users to teach the robot more efficiently as compared to the state-of-the-art approaches for learning from drawings.
In the following section, we define the key components of our problem setting.
\section{Problem Statement}
\label{sec:problem}


We consider settings where a robot learns manipulation tasks from a human teacher. We assume that the robot is equipped with a static camera that can take images of the entire task environment. Using our proposed approach, users can convey their desired task to the robot by drawing on the image of the environment (using a tablet or a similar device). Below, we identify key differences between learning from drawings and learning from real-world demonstrations. First, the human's drawings are 2-dimensional, but the robot needs to learn manipulation tasks in the 3-dimensional world. Second, in drawings, the robot cannot actually interact with objects in the environment (e.g., the robot cannot pick up a block in a drawing). Viewed together, these differences result in an information gap that our approach must resolve to successfully learn to perform tasks in the real world using the human's 2D drawings.


\p{Setup}
Before collecting the human's demonstrations, an RGB camera is placed in the environment. The camera is fixed throughout the training and evaluation process. The camera should be carefully positioned such that it can view the robot arm and its work environment, and the information loss in representing the robot states with 2-dimensional images is minimized (see Section \ref{ss:m2} for further details). Accordingly, we limit our experiments to learning manipulation tasks where the robot's movements lie within the camera’s field of view. These manipulation tasks still cover a wide array of object-centric motions like reaching, grasping, pushing, pulling, pick-and-place, pouring, scooping, etc., as well as combinations of these primitive motions.


\p{Robot}
A robot arm interacts with the environment to perform the task. 
At each step of the task, the robot reads the state of its arm $s_R \in \mathbb{R}^d$ and receives an image of the environment from our fixed camera. Consistent with the related works, we leverage language feedback from the users and vision models to extract task-relevant features from the images of the environment~\cite{zhou2022detecting,zhao2023fast}. The extracted features $o \in \mathbb{R}^k$ form the state of the environment. The overall state of the system comprises of the state of the robot and the observed features in the environment, represented as $s = (s_R, o)$. Returning to our motivating example, $s_R$ may be the position and orientation of the robot's end-effector as well as the configuration of its gripper, and $o$ could specify 
the location of the bowl the robot is trying to reach. 
When the robot performs a task in the environment, it takes an action $a \in \mathbb{R}^d$, and the state transitions according to system dynamics $\mathcal{T}(s, a)$. We do not assume that the robot has access to these transition dynamics.



\begin{figure}[t]
	\begin{center}
		\includegraphics[width=\linewidth]{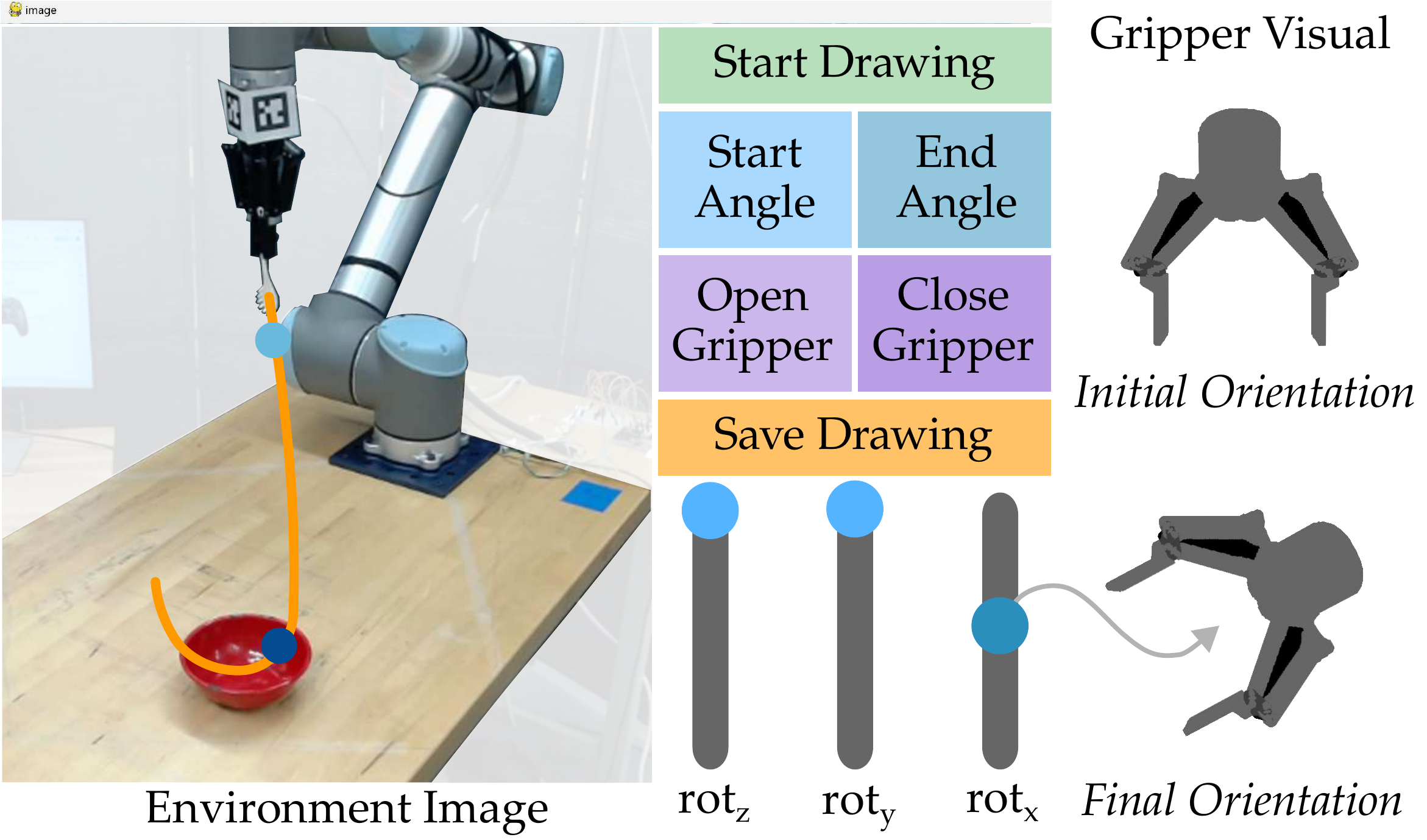}
        \vspace{-1em}
		\caption{Interface for demonstrating tasks by sketching robot trajectories. We present this interface to users on a touch-screen device. Users begin by drawing a line starting from the end-effector of the robot on the environment image shown on the left. This line
        represents the trajectory that the robot's end-effector will follow during the task. Users then specify how the end-effector should rotate by first selecting a point on the line and then selecting the orientation at that point using the $rot_x$, $rot_y$, and $rot_z$ sliders. We provide a visualization of the gripper orientation to help users identify their desired angles. In the same way, users can specify when the gripper should open or close by selecting a point on their line and then choosing the appropriate button.}
		\label{fig:interface}
	\end{center} 
\end{figure}

\p{Drawings}
The human is provided with an image of the environment taken from the fixed camera. Let $p \in \mathbb{R}^2$ be a 2D point on this image.
The human draws on this image to demonstrate their desired behavior to the robot. This drawing is provided by the human in the form of a \textit{trajectory} in the 2D space, and comprises a sequence of points 
$\xi_P = [p_1, p_2, \cdots, p_n]$,
overlaid onto the initial image of the environment (see Figure \ref{fig:interface}). 

When the human provides a drawing, they are not only thinking about the 2D trajectory (i.e., the path they draw on the image), but also about the analogous $d$-dimensional trajectory that the robot should follow in the real world (e.g., reaching the bowl and rotating the spoon). We denote this corresponding trajectory in the robot's state-space as $\xi = [s_{1}, s_{2}, \cdots, s_{n}]$. 
Moving from drawings $\xi_{P}$ to the real-world trajectories $\xi$ suffers from two fundamental challenges: 1) Since there are an infinite number of mappings from a low-dimensional to a high-dimensional space, the robot does not know \textit{a priori} how to map the trajectories drawn on the image to their real-world counterparts, and  2) the drawings do not capture how the state of the environment changes throughout the task. In other words, while the human's drawings provide information about the trajectory the robot should follow, i.e., $\xi_{R} = [s_{R_1}, s_{R_2}, \cdots, s_{R_n}]$, they do not convey how the environment state $o$ should evolve.



\p{Objective}
Given these challenges, our objective is to develop an interface that enables users to efficiently provide a diverse set of drawings that convey their desired task, and an associated algorithmic framework that can leverage these drawings to learn robot policies. 

As a first step towards this goal, we need a procedure for collecting drawings in a variety of task settings while minimizing physical interactions with the environment.
We must then convert the drawings provided by the user to real-world robot demonstrations.
These demonstrations should be in the form of state-action pairs $(s, a)$, capturing how the system state $s$ changes when the robot takes an action $a$. In other words, we want to convert the drawings $\xi_P$ provided by the human on images of the environment to a corresponding dataset of demonstrations in the real world $\mathcal{D} = \{(s_1, a_1), (s_2, a_2), \cdots, (s_n, a_n)\}$.

The robot can then leverage this dataset to learn a policy $\pi_\theta(a | s)$ that imitates the behavior that the human demonstrated in their drawings. Our proposed approach is not tied to any specific method for learning from demonstrations; but in our experiments, we leverage Behavior Cloning \cite{pomerleau1988alvinn} within our learning framework. Behavior Cloning 
matches the actions predicted by the policy in states $s$ to the corresponding actions $a$ in the dataset $\mathcal{D}$ by minimizing the following loss function to learn the policy parameters $\theta$:
\begin{equation} \label{eq:p1}
    \mathcal{L}_{BC}(\theta) = \sum_{(s, a) \in \mathcal{D}} \|\pi_\theta(s) - a\|^2
\end{equation}
Given a dataset of diverse and accurate demonstrations, we expect the robot to generalize the learned policy to new environment configurations. In the next section, we introduce our interface for obtaining diverse drawings and present our method for closing the information gap between these drawings and real-world demonstrations to ensure accurate learning.
\section{L2D2} \label{sec:L2D2}

Our aim is to make it easy for humans to teach robots by enabling them to convey the desired task through sketches. 
In particular, we want to minimize the need for physical interactions with the real world and efficiently collect diverse training data by synthetically generating new task configurations.
As detailed in the previous section, we recognize the challenges that come with operating on 2D images instead of the 3D world. 
In this section, we present our approach for addressing these issues and learning a robust robot policy.
First, in Section \ref{ss:m1}, we describe our interface for obtaining a diverse set of drawings. Next, we explain how these sketches can be accurately mapped to high-dimensional demonstrations in the real world (see Section~\ref{ss:m2}). Lastly, in Section~\ref{ss:m3}, we propose how policies learned from the diverse drawings can be grounded with a few real-world demonstrations to bridge the fundamental gap between static images and dynamic physical interactions.

\subsection{Obtaining Diverse Drawings}\label{ss:m1}

We first present our interface design and explain how users can draw on our interface to teach the robot.
To learn new tasks using imitation learning, the robot needs a dataset of state-action pairs $(s, a)$ covering a variety of task configurations.
Instead of asking users to provide these demonstrations in the real world, we use the interface shown in \fig{interface} to obtain as much of this information as possible from trajectories drawn on images of the task.

\p{Sketching Interface}
Our interface consists of three parts, each designed to convey a specific aspect of the robot's state $s_{R}$. The first part displays an image of the robot and the environment. Users draw on this image to indicate the path the end-effector should follow to perform the task. Each point $p$ on this line maps to some 3D end-effector position $p_{R}$. The second part features three sliders for changing the orientation $r$ of the end-effector about the robot's axes. Because our images are static,
it can be difficult for users to imagine how the robot rotates as they move these sliders. To make the interface more intuitive, we provide a 3D visualization of the end-effector that rotates in real-time with slider input.
The final part of the interface includes two buttons that open and close the robot's gripper $g$.

Overall, each sketch specifies a trajectory $\xi_{P}$. We update our definition of $\xi_{P}$ from Section~\ref{sec:problem} to include the end-effector rotation $r \in \mathbb{R}^{3}$ about the robot's axes and a binary gripper state $g \in \{0, 1\}$ with the 2D pixel points $p$ at each step of the trajectory. Put together,  
$\xi_P = [(p_1, r_1, g_1), (p_2, r_2, g_2), \cdots, (p_n, r_n, g_n)]$.

\begin{figure*}[t]
    \centering
    \includegraphics[width=1\linewidth]{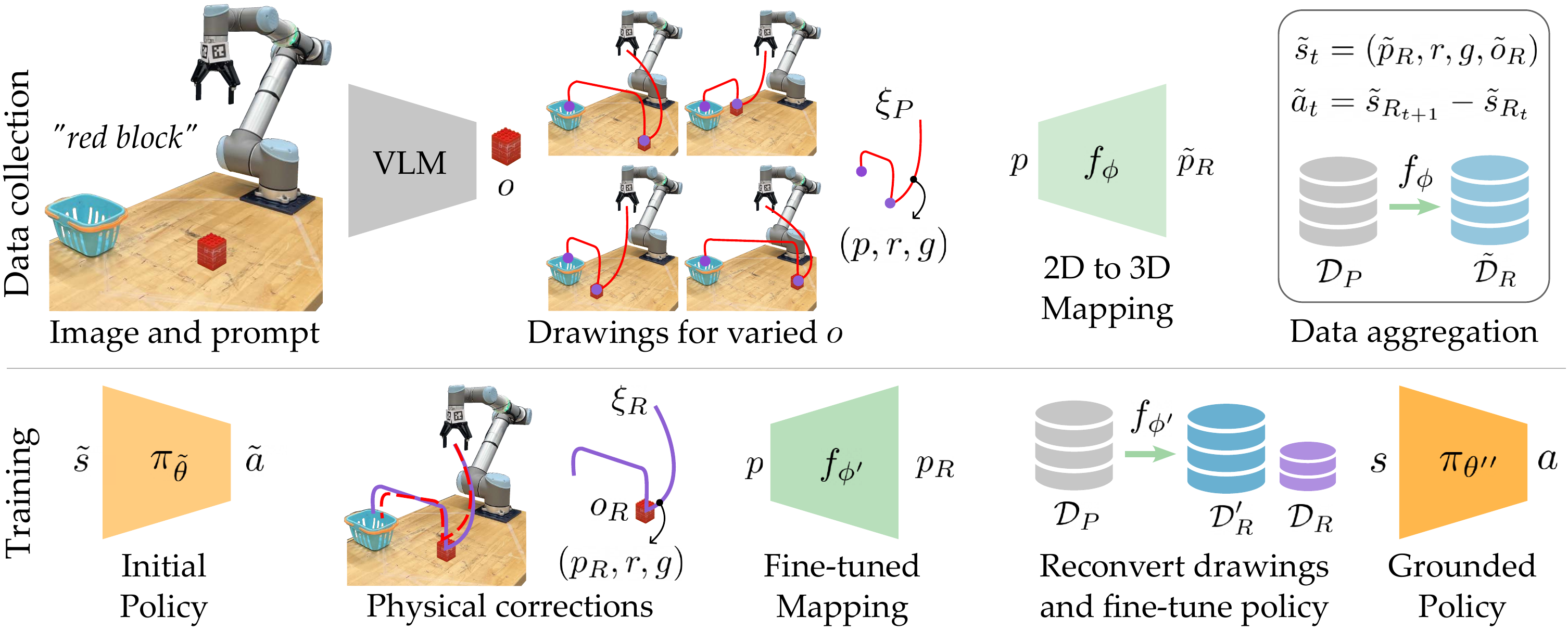}
    \caption{Proposed approach for Learning from 2D Drawings (L2D2). The top row outlines our procedure for collecting diverse sketching data. Our approach takes an initial image of the environment as input and creates multiple synthetic images covering a variety of task configurations. We achieve this by detecting relevant objects mentioned by the user using vision-language models (VLMs) and then randomly repositioning those objects in the scene. Users draw on these images to convey the desired task using our interface in \fig{interface}. 
    We then use a task-agnostic mapping to convert the 2D points in each sketch to 3D positions in the real world. 
    This ultimately results in a dataset $\tilde{\mathcal{D}}_{R}$ of state-action pairs $(\tilde{s}, \tilde{a})$ reconstructed from the drawings $\xi_{P} \in \mathcal{D}_{P}$.
    The bottom row outlines our training process. We first train a policy on the reconstructed data using behavior cloning and roll out this policy in the environment. If this policy makes any errors, users physically correct the robot's motion. These corrections result in a small dataset $\mathcal{D}_{R}$ of accurate physical demonstrations. We first use this physical data to refine our 2D-to-3D mapping and improve the quality of demonstrations reconstructed from the sketches. Then we leverage both these datasets to fine-tune the robot's policy and ground the robot's actions in the real world. Together, the diverse set of sketches and a few precise physical demonstrations result in an accurate and generalizable robot policy.}
    \label{fig:method}
\end{figure*}

With this interface, users can demonstrate different manipulation tasks. For example, in our motivating scenario of scooping cereal from a bowl, users will draw a path from a spoon held in the robot's end-effector to the center of the bowl. Then they will select points on this path where they want to rotate the end-effector and use the sliders to specify the starting and ending orientations. Our interface then linearly interpolates between these angles to capture the scooping motion.
Similarly, for the task of picking a cube and dropping it into a basket (see \fig{method}), users will draw a trajectory from the robot's gripper to the cube and then to the basket, and then select points where they want to close and open the gripper.


\p{Diverse Images}
Our interface allows users to demonstrate the task without performing it in the real world. But to learn the task effectively, the robot needs demonstrations in various task configurations (e.g., different bowl positions in the scooping task). Rather than having users physically change the environment, we leverage existing vision-language models to digitally alter the images and create new scenarios as follows.

When the robot captures an initial image of the environment, the interface asks users to specify the relevant objects through language prompts. We feed the object prompts to a vocabulary-based object detector, \textit{Detic}~\cite{zhou2022detecting} that extracts object locations $o$ in the image, and segments out the object masks. We use these masks to generate new environment images by changing their position in the image and inpainting the area where the objects were moved from (see \fig{method}).


In our experiments, we found that moving objects to random image locations introduces sufficient variety for learning the task. For each generated image, users provide a drawing as described earlier. These drawings are related because they show the same task, but each drawing is unique since the way that task is performed changes. This way, the robot can efficiently obtain a diverse dataset of $m$ trajectories $\mathcal{D}_{P} = \{\xi_{P_1}, \xi_{P_2}, \ldots, \xi_{P_m}\}$ without requiring users to physically interact with the robot or environment.

\subsection{Converting Drawings to Robot Trajectories} \label{ss:m2}


Now that we have a dataset of drawings $\mathcal{D}_{P}$, we need to translate these drawings into state-action pairs that the robot can use to learn a control policy. 
We know that each drawing $\xi_{P}$ corresponds to a real-world robot trajectory $\xi$ that the user has in mind. However, when the user projects the trajectory from a high-dimensional robot state space into a low-dimensional image space, we inevitably lose some information. 
In this section, we theoretically quantify this information loss, and propose a solution for mapping the 2D pixel points back to the 3D world with minimal reconstruction error.

\p{Information Loss} 
We start by formalizing the fundamental gap between drawings and real robot states. In our motivating example, the robot state $s_{R}$ includes the end-effector position $p_{R}$, orientation $r$, and gripper state $g$. Our drawings provide direct rotation $r$ and gripper $g$ information, but when we take an image of the environment, the 3D robot positions $p_{R} = [x_{R}, y_{R}, z_{R}]$ are reduced to 2D pixel points $p = [x_{p}, y_{p}]$ on the image plane. To model this projection, we first map the states to the camera frame $C$:
\begin{equation}\label{eq:m2}
    [p_{C} \;\; 1]^{T} = T_{CR} \cdot [p_{R} \;\; 1]^{T}
\end{equation}
Here $T_{CR}$ is a homogeneous transformation from the robot's reference frame $R$ to the camera frame $C$. Then, we project the states $p_{C} = [x_{C}, y_{C}, z_{C}]$ onto the image by dropping the $z$ coordinate, and scaling the $x{-}y$ coordinates based on the dropped $z_{C}$ (which is their depth) and the camera's focal lengths $(f_{x}, f_{y})$ to account for perspective~\cite{hartley2003multiple}.
\begin{equation}\label{eq:m3}
    p = \left[x_{C}\frac{f_{x}}{z_{C}}, \; y_{C}\frac{f_{y}}{z_{C}} \right]
\end{equation}

The transformation and scaling steps preserve all information --- we can retrieve the robot states $s_{R}$ given their distance $z_{C}$ along the camera's $z$-axis. Therefore, we only lose information when we discard $z_{C}$ to project the states onto the image.
Note that a depth camera would not be helpful in this case, since users still provide a drawing in 2D. 
For example, the user can show how the robot moves left or right in the image plane, but not how far \textit{into the frame} the robot should go.

We quantify this loss as the proportion of the total variance explained by the $z$-coordinate of all robot states $p_{C} \in P_{C}$ expressed in the camera frame~\cite{jolliffe2016principal}:
\begin{equation}\label{eq:var_loss}
    I_{loss}(C) = \frac{Var(z_{C})}{Var(x_{C}) + Var(y_{C}) + Var(z_{C})}
\end{equation}
The more the robot states vary along the direction that the camera faces, the higher the information loss when representing these states with a 2D sketch.

\p{Camera Placement}
We want to minimize this information loss so that the robot can accurately reconstruct the states from the points drawn by users on the camera image.
Prior approaches address this gap by obtaining additional information with each drawing in the form of distance inputs~\cite{gu2023rt} (i.e., users need to manually specify the depth or height) or complementary sketches on images taken from orthogonal viewpoints~\cite{yu2025sketch} (i.e., users draw the same task from multiple perspectives), both of which increase the user's burden. 
Instead of seeking extra inputs, we propose leveraging our analysis in \eq{var_loss} to find a new camera placement that minimizes the variance along the $z$-axis of the image plane:
\begin{equation}
    C^{*} = \argmin_{C} \; I_{loss}(C)
\end{equation}

We can derive the optimal solution for the above objective using Principal Component Analysis (PCA)~\cite{jolliffe2016principal}. In PCA, we compute the directions of maximum variance (i.e., the principal components) as the eigenvectors of the covariance matrix $\Sigma_{C}$ of robot states. 
\begin{equation}\label{eq:m5}
    \Sigma_{C} = \frac{P_{C}P_{C}^{T}}{n-1}
\end{equation}
Here $P_{C}$ is a matrix of uniformly sampled robot states represented in the camera frame. Let $\Lambda = [\lambda_{1}, \lambda_{2}, \lambda_{3}]$ be the eigenvalues of the covariance matrix arranged in decreasing order of magnitude and $V = [v_{1}, v_{2}, v_{3}]$ be an orthonormal matrix of corresponding eigenvectors. The first principal component $v_{1}$ represents the direction of maximum variance in robot states, while the last component $v_{3}$ captures the least variance. To minimize the reconstruction error, we want the $z$-axis of the camera to be aligned with the last eigenvector and the $x{-}y$ image plane to be parallel to the plane formed by the first two principal components. This arrangement will result in the least information loss $I_{loss}^{*}$ given by:
\begin{equation}
    I_{loss}(C^{*}) = \frac{\lambda_{3}}{\sum_{i=1}^{3} \lambda_{i}}
\end{equation}

Our analysis thus far quantifies the data loss for projecting 3D robot states to 2D pixel points and proposes mitigating this loss through an improved camera setup. We do not assume any prior knowledge of the tasks that the human wants to teach, and consider the entire robot state space in our calculations. However, when performing tasks in the real world, the robot may operate in a more restricted subset of states $\mathcal{W} \subset P_{C}$. 
For instance, the robot's workspace may be confined by an adjacent wall or be limited to a table that it is mounted on.
In practice, we leverage knowledge of this \textit{working region} $\mathcal{W}$ to compute a more domain-specific covariance matrix using $\mathcal{W}$ instead of $P_{C}$ in \eq{m5} and further improve the camera position.

With the camera held fixed, we now shift our focus to how the robot can accurately map the points $p$ drawn on the image to the corresponding robot positions $p_{R}$.

\p{2D to 3D Mapping}
According to our camera model, the 3D robot positions are linearly projected on the 2D image plane, followed by a nonlinear scaling. 
When retrieving the positions from the pixel points, we can combine all inverse transformations into a single mapping $f: p \rightarrow p_{R}$.
Prior work has shown that non-linear mappings can encode and extract information more effectively than PCA projections~\cite{bahadur2022dimension,rolinek2019variational,fournier2019empirical,cacciarelli2023hidden}.
Building on this insight, we model $f$ as a non-linear function represented by a neural network $f_\phi(p)$ with parameters $\phi$. 
We train this network on a new calibration dataset $\mathcal{D}_{map} = \{(p_1, p_{R_1}), (p_2, p_{R_2}), \ldots \}$, where each robot position $p_{R_i}$ is uniformly sampled from the space $\mathcal{W}$ and projected onto the image using \eq{m2} and \eq{m3} to get the matching pixel point $p_i$.
Note that this data is \textit{task-agnostic} and is automatically generated before collecting the task-specific user drawings.

We minimize the following reconstruction loss function to learn the mapping from 2D points to 3D robot states using $\mathcal{D}_{map}$:
\begin{equation} \label{eq:m8}
    \mathcal{L}_{map}(\phi) = \sum_{(p, p_{R}) \in \mathcal{D}_{map}} \mid\mid f_\phi(p) - p_{R} \mid\mid^2
\end{equation}

We expect the mapping $f_{\phi}$ trained on the $\mathcal{L}_{map}$ to find non-linear manifolds that better fit the state distribution than linear principal components, leading to more accurate reconstructions. In practice, this can help us further reduce the information loss~\cite{fournier2019empirical}, such that:
\begin{equation}
    I_{loss}^{nonlinear}(C) \leq I_{loss}(C)
\end{equation}

Algorithm~\ref{alg:m1} summarizes our proposed approach for bridging the gap between 2D paths drawn on images and corresponding robot trajectories in the 3D world.
In the next part of our approach, we describe how we apply the learned mapping $f_{\phi}$ to convert the sketch data $\mathcal{D}_{P}$ collected by our interface into real-world demonstrations that can be used to train the robot.

\begin{algorithm}[t]
    \caption{2D to 3D Mapping}
    \label{alg:m1}
    \begin{algorithmic}[1] 
        \State Given: Workspace $\mathcal{W}$, Camera positions $\mathcal{C}$
        
        \State Find camera placement $C^* = \min_{C\in \mathcal{C}} I_{loss}(C)$
        \State Initialize $\mathcal{D}_{map} = \{\}$
        \For {position $p_{R}\sim\mathcal{W}$}
            \State Get pixel point $p$ with Eq.~\ref{eq:m2} and Eq.~\ref{eq:m3}
            \State $\mathcal{D}_{map} \cup (p, p_{R})$
        \EndFor
        \State Initialize $f_{\phi}$ with random $\phi$
        \For {$(p, p_{R})$ in $\mathcal{D}_{map}$}
            \State Compute loss $\mathcal{L}_{map}(\phi)$ using Eq. \ref{eq:m8}
            \State Update weights $\phi$ to minimize $\mathcal{L}_{map}$
        \EndFor
        \State \Return Reconstruction function $f_\phi$
        
    \end{algorithmic}
\end{algorithm}

\subsection{Learning from Drawings}\label{ss:m3}

Following our outline in \fig{method}, we have obtained a diverse dataset of sketches $\mathcal{D}_{P}$ with our interface and learned a task-agnostic function $f_\phi(p)$ that maps 2D image points $p$ to 3D robot positions $p_{R}$ with minimal information loss.
To train the robot policy, we now need to convert the drawings into a dataset of state-action pairs $(s,a)$ that capture the desired behavior.
In what follows, we describe our procedure for extracting this data using the learned mapping $f_{\phi}$, and then present our core idea of grounding the drawing data with a few real-world demonstrations to fill in any remaining gaps in information and learn robust robot policies.

\p{Data Aggregation}
We begin by aggregating the data collected by our interface.
Here we apply the mapping learned in Section~\ref{ss:m2} to convert each drawing $\xi_{P} \in \mathcal{D}_{P}$ to a sequence of robot states $\tilde{\xi}_{R} = [(\tilde{p}_{R_{1}}, r_{1}, g_{1}), (\tilde{p}_{R_{2}}, r_{2},\\ g_{2}), \ldots, (\tilde{p}_{R_{n}}, r_{n}, g_{n})]$ by reconstructing the positions. 
$$f_{\phi}(p) = \tilde{p}_{R}$$
The accent $\tilde{\;\;\;}$ denotes the approximate reconstruction we get when moving from a 2D to a 3D space. 
The tuple $(\tilde{p}_{R}, r, g)$ defines the robot's state $\tilde{s}_{R}$, but it does not include the environment state $o_{R}$. In general, the user's drawings do not provide any information about other objects in the environment; the robot only observes the initial environment image and does not see how it will evolve when rolling out the drawn trajectory.
For example, in \fig{method}, users do not sketch the cube's motion.
Without this feedback, the robot will go to the basket even if it fails to grasp the cube, since it does not know that the cube should move with its gripper.


One way we can address this problem is by simulating how the object state evolves throughout the task. In our implementation, we approximate object dynamics based on the object locations $o$ detected in the initial image of the environment (from Section~\ref{ss:m1}) and the reconstructed trajectory $\tilde{\xi}_{R}$ using the following rule:
First we leverage our learned mapping $f_{\phi}$ to convert the pixel locations $o$ to 3D object positions $\tilde{o}_{R} = f_{\phi}(o)$. Then, to model how the objects move during the task, we change their positions along with the end-effector positions $\tilde{p}_{R}$ whenever the gripper $g$ is closed within a pre-specified distance from the objects. 
While we apply this practical simplification, it is not central to our approach; practitioners may instead employ physics-induced image manipulation~\cite{zheng2012interactive} or video generation from task descriptions~\cite{bharadhwaj2024gen2act} to simulate object dynamics. 

Once we have the simulated object positions $\tilde{o}_{R}$, we combine them with the reconstructed robot state $\tilde{s}_{R}$ to get the complete task state $\tilde{s} = (\tilde{s}_{R}, \tilde{o}_{R})$. 
To create the training dataset, we now need to know what actions the robot should take in these states to achieve the sketched task in the real world. We do that by setting the robot's actions to be the difference between consecutive robot states in the reconstructed trajectory: 
$$\tilde{a}_{t} = \tilde{s}_{R_{t+1}} - \tilde{s}_{R_{t}}$$
With this final piece of information, we can construct the training dataset $\tilde{\mathcal{D}}_{R} = \{(\tilde{s}_{1}, \tilde{a}_{1}), (\tilde{s}_{2}, \tilde{a}_{2}), \ldots \}$. 
Despite our efforts in accurately reconstructing the robot state and simulating object dynamics, this data can be imperfect due to the inherent lack of information in low-dimensional sketches and static images.
That said, while this data lacks in accuracy, it compensates for it with a rich variety of task configurations.

In the final part of our approach, 
we present our idea for improving the accuracy of this diverse dataset and training a robot policy that can account for dynamic interactions that are absent in our drawings.

\p{Grounding with Physical Demonstrations}
We use the reconstructed data to train a preliminary robot policy
$\pi_\theta(s) \to a$ that maps task states to robot actions.
Let $\pi_{\tilde{\theta}}$ be the policy trained on $\tilde{\mathcal{D}}_{R}$ using the behavior cloning objective in Equation \ref{eq:p1}.
In tasks involving simple physical interactions, this initial policy alone may be sufficient for successfully completing the task.

However, when learning contact-rich tasks, we enable humans to refine the robot's behavior with a small dataset of accurate real-world demonstrations $\mathcal{D}_{R} = \{(s_1, a_1), (s_2, a_2), \cdots \}$.
This data can be collected by executing the task according to $\pi_{\tilde{\theta}}$ and asking users to correct the robot's trajectory, or by obtaining new demonstrations through user teleoperation. 
While $\mathcal{D}_{R}$ may not cover as many scenarios as in $\tilde{\mathcal{D}}_{R}$, it precisely grounds the synthetic observations in the real world. 
Our insight is that these data sources are complementary. 
Below, we describe how we can best utilize both datasets to train an accurate and generalizable policy.

Our approach does not simply merge the two sets like DAgger~\cite{ross2011reduction}. Rather, we utilize the real-world demo-nstrations to ground the robot policy in two steps: (i) recalibrating the data derived from user drawings, and then (ii) refining the robot policy using both the real-world and drawing datasets. 
In the first step, we recall that our 2D to 3D mapping $f_{\phi}$ was trained on positions uniformly sampled from a task-agnostic space $\mathcal{W}$ (see Section~\ref{ss:m2}). 
By contrast, $\mathcal{D}_{R}$ only includes states that are specific to the desired task, which may or may not be represented in $\mathcal{W}$.
To address this mismatch, we adapt our mapping to the user's task by creating new training pairs $(p, p_{R})$ with real robot positions in $\mathcal{D}_{R}$. With these task-specific pairs, we fine-tune our pre-trained mapping $f_{\phi}$. Specifically, we update its weights $\phi$ with gradients from the loss in \eq{m8} computed over the new states:
\begin{equation}
    \phi' \leftarrow \phi - \alpha \nabla_{\phi} \mathcal{L}_{map} \quad\; \text{using  } (p, p_{R}) \sim \mathcal{D}_{R}
\end{equation}
The updated mapping $f_{\phi'}$ captures task-specific information and should thus provide a more precise 2D to 3D mapping for the task at hand. 
Now that we have this improved mapping, we reapply it to the original user drawings $\mathcal{D}_{P}$ to construct a more accurate dataset of reconstructed trajectories $\mathcal{D}_{R}'$.


In the second step, we leverage this new dataset of demonstrations $\mathcal{D}_{R}'$ extracted from drawings and the physical demonstration data $\mathcal{D}_{R}$ to retrain the robot policy.
Here we highlight that besides task-specific state information, $\mathcal{D}_{R}$ also contains rich object interactions.
For instance, users may demonstrate grasping the cube multiple times if they fail during initial tries. 
These interactions offer valuable feedback that can help the robot adapt to real-world outcomes, e.g., learning not to lift the cube until it is successfully grasped.

To incorporate this knowledge in training the robot's policy, one may simply try to combine the two datasets. But in practice, due to the imbalance between the sizes of the real-world and drawing data, the learned policy becomes biased towards the trajectories in $\mathcal{D}_{R}'$ and ignores the contact-rich information in $\mathcal{D}_{R}$.
In our implementation, we mitigate this bias by sequentially training the robot's policy on each dataset using the standard behavior cloning loss from \eq{p1}:
\begin{equation}
\begin{aligned}\label{eq:m11}
    &\theta' \gets \theta - \beta \nabla_{\theta} \mathcal{L}_{BC} \quad\quad \text{using  } (s', a') \in \mathcal{D}_{R}'\\
    &\theta'' \gets \theta' - \beta \nabla_{\theta'} \mathcal{L}_{BC} \quad\; \text{using  } (s, a) \in \mathcal{D}_{R}
\end{aligned}
\end{equation}
Here $\theta'$ denotes the intermediate policy parameters after training on $\mathcal{D}_{R}'$. These parameters capture behaviors across various task settings. We then fine-tune $\theta'$ on $\mathcal{D}_{R}$ to obtain the final grounded policy $\pi_{\theta''}$. This final training phase helps the robot extrapolate the real-world knowledge to the diverse settings illustrated in our drawings.

Overall, our training procedure enables the robot to not only adapt to different task configurations --- because of the diverse drawings collected by our interface --- but also realize the physical consequences of its actions and compensate for any inaccuracies in 3D reconstruction and simulation of object dynamics.

\begin{algorithm}[t]
    \caption{$L2D2$}
    \label{alg:m2}
    \begin{algorithmic}[1] 
        \State Given: Drawings $\mathcal{D}_{P}$, Mapping $f_{\phi}$
        \vspace{0.5em}
        \State Initialize $\tilde{\mathcal{D}}_{R} = \{\}$
        \Function {$LearnFromDraw$}{$\mathcal{D}_P, f_\phi$}
        \For {$\xi_{P}$ in $\mathcal{D}_{P}$}
            \State // \textcolor{orange}{\textit{Data Aggregation from Drawings}}
            \For {$(p_{t}, r_{t}, g_{t})$ and $p_{t+1}$ in $\xi_{P}$}
                \State Convert $\tilde{p}_{R_{t}} \gets f_{\phi}(p_{t})$ 
                \State Convert $\tilde{p}_{R_{t+1}} \gets f_{\phi}(p_{t+1})$ 
                \State Robot state $\tilde{s}_{R} = (\tilde{p}_{R}, r, g)$
                \State Compute action $\tilde{a}_{t} = \tilde{s}_{R_{t+1}} - \tilde{s}_{R_{t}}$
                \State Get object state $\tilde{o}_{R}$ from image
                \State $\tilde{\mathcal{D}}_{R} \cup (\tilde{s}_{R}, \tilde{o}_{R}, \tilde{a}_{t})$
            \EndFor
        \EndFor
        \vspace{0.5em}
        \State Initialize $\pi_{\theta}$ with random $\theta$
        \For {$(s, a)$ in $\tilde{\mathcal{D}}_{R}$}
            \State Compute loss $\mathcal{L}_{BC}(\theta)$ using Eq. \ref{eq:p1}
            \State Update weights $\theta$ to minimize $\mathcal{L}_{BC}$
        \EndFor
        \vspace{0.5em}
        
        \State \Return Robot policy $\pi_{\theta}$
        \EndFunction

        \vspace{0.5em}

        \State // \textcolor{orange}{\textit{Learning initial robot policy from drawings}}
        \State $\pi_{\tilde{\theta}} \gets \text{$LearnFromDraw$}(\mathcal{D}_{P}, f_{\phi})$
        \vspace{0.5em}
        \State Collect real-world demonstrations $\mathcal{D}_R$ 
        \vspace{0.5em}
        
        \State // \textcolor{orange}{\textit{Fine-tuning the reconstruction function}}
        
        \For {$(p, p_{R})$ in $\mathcal{D}_{R}$}
            \State Compute loss $\mathcal{L}_{map}(\phi)$ using Eq. \ref{eq:m8}
            \State Update weights $\phi$ to minimize $\mathcal{L}_{map}$
        \EndFor
        \State Task-specific mapping $f_{\phi'}$
        \vspace{0.5em}
        \State // \textcolor{orange}{\textit{Learning intermediate policy parameters}}
        \State $\pi_{\theta'} \gets \text{$LearnFromDraw$}(\mathcal{D}_{P}, f_{\phi'})$
        \vspace{0.5em}
        \State // \textcolor{orange}{\textit{Fine-tuning robot policy}}
        \For {$(s, a)$ in $\mathcal{D}_{R}$}
            \State Compute loss $\mathcal{L}_{BC}(\theta')$ using Eq. \ref{eq:p1}
            \State Update weights $\theta'$ to minimize $\mathcal{L}_{BC}$
        \EndFor
        \vspace{0.5em}
        \State \Return Grounded robot policy $\pi_{\theta''}$

    \end{algorithmic}
\end{algorithm}

\subsection{Algorithm Summary}
Our proposed approach for \textbf{L}earning from \textbf{2D D}rawings: L2D2, is summarized in Algorithms \ref{alg:m1} and \ref{alg:m2}. 
Our code is available here: \url{https://github.com/VT-Collab/L2D2}

We first apply PCA to find an optimal camera placement $C^{*}$ for taking images of the environment, and learn a 2D to 3D mapping $f_{\phi}$ following the lines $8{-}12$ in Algorithm~\ref{alg:m1}.
This mapping reconstructs the real-world positions corresponding to the points in images taken from $C^*$.
We then leverage vision-language models~\cite{zhou2022detecting} to synthetically generate images of varying task configurations and ask users to convey their desired task by drawing on these images using our interface in \fig{interface}. This creates a diverse dataset of drawings $\mathcal{D}_{P}$.

To learn from these drawings, we use our mapping $f_{\phi}$ to create a dataset of reconstructed state-action pairs $(\tilde{s}, \tilde{a}) \in \tilde{\mathcal{D}}_{R}$ as outlined in Algorithm~\ref{alg:m2}.
With this data, we train an initial robot policy $\pi_{\tilde{\theta}}$.
Since this policy is learned from static drawings, it may fail to capture dynamic real-world interactions. 
We bridge this gap by grounding the policy with a few real-world demonstrations $\mathcal{D}_{R}$.
This process involves two steps: First, we improve our mapping $f_{\phi'}$ using task-specific states from the real-world data (lines $26{-}30$) and reconstruct the drawing demonstrations $\mathcal{D}_{R}'$. 
Second, we leverage both $\mathcal{D}_{R}'$ and $\mathcal{D}_{R}$ to sequentially train a grounded robot policy $\pi_{\theta''}$ (lines $32{-}38$).
The diverse drawings enable the robot to generalize across various task configurations, while the real-world data teaches it to account for environment dynamics. 

Although ours is not the first approach for learning from sketches, it introduces two key advances: synthetically generating diverse drawings and grounding them with real-world demonstrations.
In the following section, we evaluate how these advances enable humans to teach robots more efficiently and accurately than prior sketch-based learning approaches.

\section{Real-World Experiments} \label{sec:experiments}
In Section~\ref{sec:L2D2}, 
we presented L2D2, an interface-based approach for teaching robots with task sketches.
We now evaluate our proposed approach through real-world experiments. Specifically, we compare L2D2 to state-of-the-art methods for learning from drawings, as well as standard approaches for learning from physical demonstrations.
We break down our experiments into three parts. First in Section \ref{ss:e1}, we evaluate the performance of L2D2 on short manipulation tasks with data provided by expert users. Next, in Section \ref{ss:e2}, we conduct an in-person study to assess whether novice users can leverage our approach to teach robots efficiently. Finally, in Section \ref{ss:e3}, we test if expert users can apply our approach for teaching robots to perform longer manipulation tasks.

\p{Independent Variables}
We compare L2D2 to three state-of-the-art baselines that leverage different feedback mechanisms for teaching manipulation tasks to a robot arm: using teleoperated demonstrations (\textbf{Teleop}), using sketches to condition the robot policy (\textbf{RT-Traj}) \cite{gu2023rt}, and using sketches in two camera images to reconstruct the 3D demonstrations (\textbf{S2S}) \cite{yu2025sketch}.
We also compare against two ablations of our proposed approach. In the first ablation, we evaluate the performance of L2D2 when it only has access to drawings collected using our interface (\textbf{L2D2-D}). In the second ablation, we evaluate the performance of our approach when it is trained with just the small set of physical corrections that we collect to ground the drawings (\textbf{Teleop-min}). Below, we describe these approaches and their training and operating procedures in detail:
\begin{itemize}
    \item \textbf{Teleop:} In Teleop, users directly control the position and orientation of the robot's end-effector, and actuate the robot's gripper using a joystick. 
    With this approach, users can provide physical demonstrations that accurately capture their desired behavior in the real-world environment.
    
    \item \textbf{RT-Traj:} 
    In this approach, users draw the robot's trajectory on a camera image. 
    Unlike L2D2, RT-Traj does not get rotational inputs, instead, it asks users to specify the real-world heights for key points along the sketched trajectory.
    This sketch is used to condition a pre-trained transformer that outputs robot actions. So for each sketch, we roll out these actions to record the corresponding physical demonstration (that we use to train the robot policy).

    The transformer model used to map the 2D sketches to the 3D world is trained on large multi-task data.
    To reduce this data requirement, we simplify its implementation by directly giving it the sketched trajectories rather than images of the sketch as in the original paper~\cite{gu2023rt}.
    In our experiments, we collect $160$ task-specific demonstrations to train this mapping. These demonstrations are different from those used to train the robot policy.
    
    
    \item \textbf{S2S:} 
    This is another sketching approach that takes images of the environment, but from two orthogonally placed cameras instead of one. 
    Users draw on both camera images to convey a single corresponding real-world demonstration. So users need to imagine how the same robot trajectory would appear from two different viewpoints.
    Similar to RT-Traj, S2S does not obtain rotational inputs from users and requires a pre-trained autoencoder network to convert the paired 2D sketches into one real-world trajectory. We use the same $160$ task-specific demonstrations that we collected for RT-Traj to train the mapping for S2S following the procedure in \cite{gu2023rt}.
    However, unlike RT-Traj, S2S directly maps the sketches to real-world demonstrations without the need for rolling out the sketches in the real world.
    
    
\end{itemize}

For all baseline methods, users need to physically interact with the environment to vary the task scenarios, which they do not need to do with L2D2.
We also do not require large task-specific datasets to learn a 2D to 3D mapping.
Instead, we use a task-agnostic mapping $f_{\phi}$ and a few real-world demonstrations to ground the diverse drawings collected by our interface.

\p{Experimental Setup}
In all our experiments, we use a $6$-DoF Universal Robots UR-10 robot arm equipped with a two-finger Robotiq gripper. All the tasks are performed on the table on which the robot arm is mounted. We leverage the procedure in Section~\ref{ss:m1} to position the camera as best as we can to minimize the information loss when learning from 2D drawings. The same camera position is used to collect drawings for all sketch-based methods and all the tasks in the experiments. In addition to this camera, we place a second camera almost orthogonal to the first one for the S2S baseline.

During each interaction, the user first resets the environment by physically changing the positions of task-relevant objects. This happens automatically before each drawing for L2D2 and L2D2-D.
Then they provide a sketch or a teleoperated demonstration in the current scenario to convey their task to the robot. 

\subsection{Short Horizon Tasks} \label{ss:e1}

We start by evaluating the performance of all methods in short-horizon manipulation tasks like lifting or pushing objects placed on a table.
Here we provide the same amount of expert demonstrations (sketched or physical) to each approach and test how well the resulting policy performs the demonstrated task.

\begin{figure}
    \centering
    \includegraphics[width=1.0\linewidth]{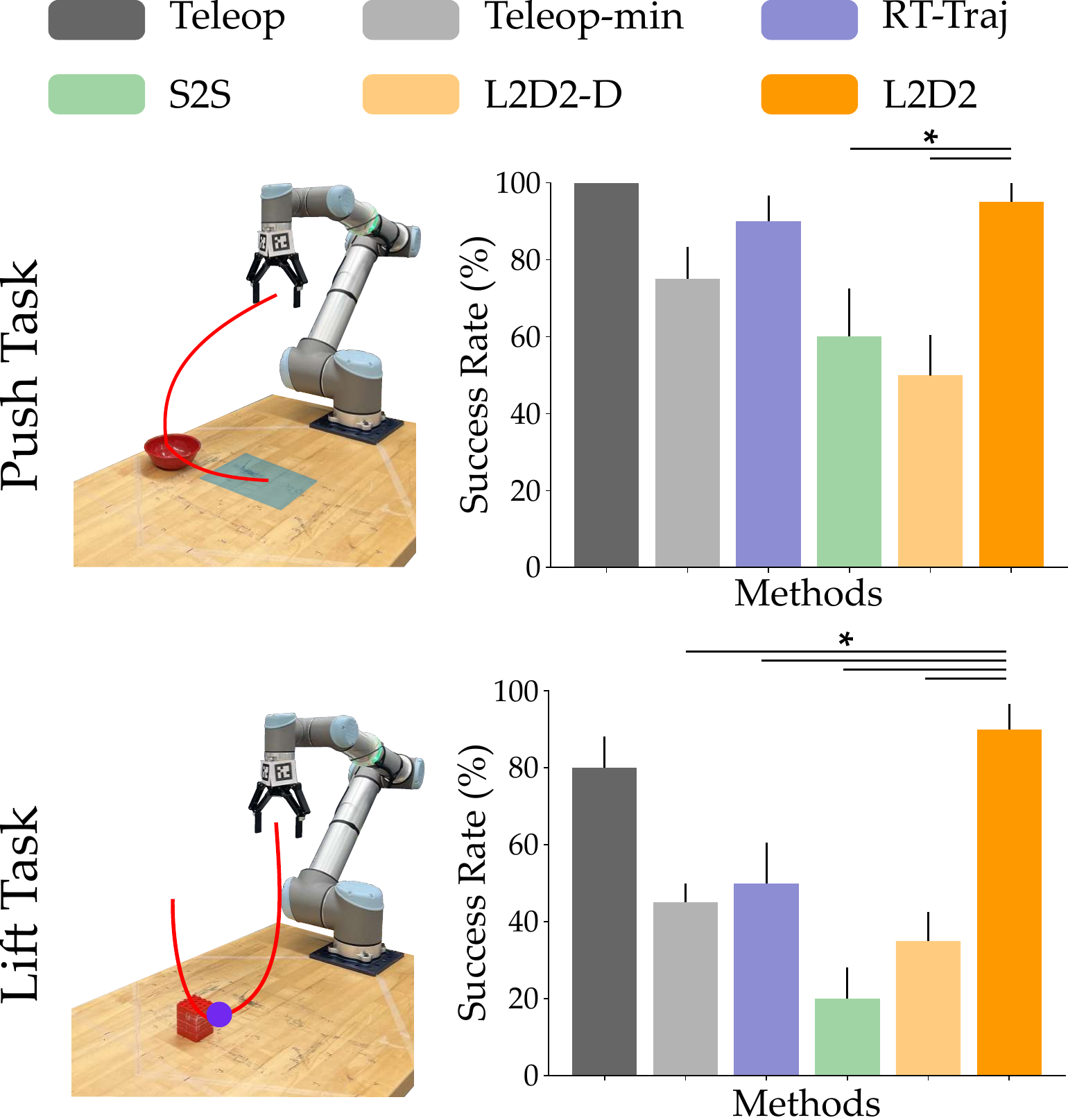}
    \caption{Results for short-horizon tasks with expert data. 
    (Top) The user is trying to teach the robot to push the bowl to the center of the table (blue region) by drawing the trajectory for the task on the image of the environment. (Bottom) The robot is learning to pick up a block from the drawings provided by the user. We report the success rate for both tasks averaged over $10$ independent rollouts with varying object locations. The error bars show the standard error around the mean (SEM), and $*$ denotes statistical significance $(p<0.05)$. For the push task, L2D2 achieves a higher success rate than L2D2-D and S2S, while for the \textit{Lift} task L2D2 performs similar to Teleop and outperforms all other baselines.}
    \label{fig:exp1}
\end{figure}

\p{Task Descriptions}
In this experiment, expert users taught two tasks to the robot arm: \textit{Lift} and \textit{Push} (shown in Figure \ref{fig:exp1}). In the \textit{Lift} task, the expert's goal was to teach the robot to reach a red cube placed on the table, close its gripper, and lift the object above a specified height threshold. In the \textit{Push} task, the expert had to teach the robot to reach for a bowl and move it to the center of the table.
For the \textit{Lift} task, the experts were instructed to randomly vary the cube's position to uniformly cover the entire table, and for the \textit{Push} task, they were asked to ensure that the random bowl placement was away from the table's center.

\p{Data Collection and Training}
We fix the total number of demonstrations that the experts provide for each method. 
Specifically, they provided a total of $60$ demonstrations to the robot per task. When using Teleop, the experts gave $60$ physical demonstrations. For RT-Traj, they drew $60$ sketches, while for S2S, they made $120$ drawings (one in each camera frame per demonstration). Finally, for L2D2, the experts provided $50$ drawings and $10$ physical demonstrations. 

Once we have the data for all methods, we use the same network architecture to train their respective policies. Specifically, we use a multi-layer perceptron (MLP) that takes the end-effector state and object features as input and outputs the end-effector velocities.

\p{Dependent Variables}
We evaluate the performance of each approach in terms of success rate for completing $10$ different instances (i.e., initial object positions) of each task. In each instance, we compute success by breaking down the task into smaller segments and measuring the fraction of segments that were completed successfully. For the \textit{Lift} task, we divide the task into reaching and lifting segments to measure success. For example, if the robot reaches the block but fails to grasp and lift it, we have $50\%$ success. Whereas if the robot successfully grasps and lifts the block, the success is $100\%$.
Similarly, we divide the \textit{Push} task into two segments: reaching the bowl and pushing it to the center, each accounting for $50\%$ of the task success.


\p{Results}
Figure \ref{fig:exp1} summarizes our results averaged over the $10$ testing scenarios.

We first analyze the success rates for the \textit{Push} task. A One-way ANOVA revealed that the teaching method had a significant effect on task success ($F(5, 54) = 6.02$ and $p<0.05$). 
Post-hoc comparisons
indicated that L2D2 significantly outperformed S2S ($p<0.05$) while being as successful as Teleop ($p=0.67$) and RT-Traj ($p=0.67$).
We found similar results for the \textit{Lift} task. A One-Way ANOVA revealed that the choice of the teaching method had a significant effect on the robot's performance ($F(5, 54) = 11.55$ and $p<0.05$). Post-hoc analysis indicated that L2D2 significantly outperformed both S2S and RT-Traj baselines ($p<0.05$),  while performing similarly to Teleop ($p=0.37$). 


This result shows that despite using only $10$ real-world demonstrations, L2D2 was as performant as Teleop, which has access to $60$ physical demonstrations.  
In contrast, S2S failed to reach the block accurately in many instances. We hypothesize that this is due to the difficulty of visualizing how the robot's trajectory would appear in two different camera frames. Our results for S2S match those reported in \cite{yu2025sketch} for the tasks involving gripper actuation.

When comparing to ablations of our approach, we found that L2D2 achieves a significantly higher success rate than L2D2-D in both tasks ($p<0.05$). It also outperforms Teleop-min in the \textit{Lift} task ($p<0.05$) where there is a greater variation in object positions. 
This highlights the benefit of grounding the drawings with a few physical demonstrations, while also showing how the diverse drawings can help the robot perform better under varying task configurations.

\subsection{User Study} \label{ss:e2}
In the previous section, we demonstrated that our proposed approach can learn short manipulation tasks effectively with drawings from expert users.
These users were experienced in using the sketching interfaces and were thus able to provide high-quality drawings to the robot.
However, can novice users learn to use our interface effectively, and will they prefer our proposed approach for teaching robots?
Along with improving performance, it is equally important for a teaching interface to be intuitive for human teachers.
Hence, in this section, 
we conduct a user study with $12$ participants who had never used the drawing interfaces and assess whether it is easy for these end-users to teach the robot by drawing on our interface. We evaluate the time they need to provide drawings and the resulting performance of robot policies learned from their data.

\begin{figure*}
    \centering
    \includegraphics[width=1.0\linewidth]{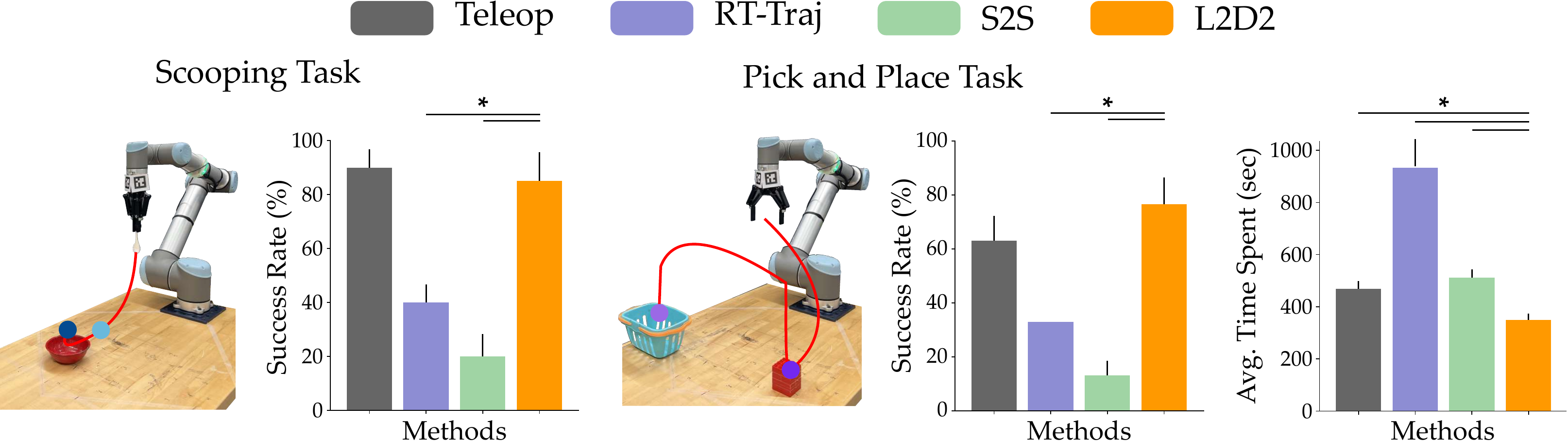}
    \caption{Objective results for the user study in Section \ref{ss:e2}. Participants teach the robot to perform two tasks in the environment: \textit{Scooping} and \textit{Pick and Place}. In each task, they provide physical demonstrations through teleoperation (Teleop) and drawings using our proposed approach (L2D2) and two sketching baselines, RT-Traj and S2S. We record the total time spent by the users in providing demonstrations to the robot and the average success rate of the learned policy evaluated over 10 rollouts. The error bars in the plots show the SEM and $*$ signifies that L2D2 had a significantly better performance than the baseline. Across both tasks, users spend significantly less time providing demonstrations using L2D2, and achieved a significantly higher success rate as compared to RT-Traj and S2S.}
    \label{fig:exp2}
\end{figure*}

\p{Task Descriptions}
In this study, the users were tasked with teaching two tasks: \textit{Pick and Place} and \textit{Scooping}. In the \textit{Pick and Place} task, the users' goal was to teach the robot to pick up a block from different locations in the environment and drop it inside a bin kept at a fixed location. For the \textit{Scooping} task, the robot started with a spoon in its gripper. The users' goal was to teach the robot to reach inside the bowl and rotate the robot's gripper such that the spoon scoops the contents of the bowl. For both tasks, users were asked to randomly place the objects in the environment at the start of each interaction, except when using L2D2.

\p{Participants and Procedure}
We recruited $12$ participants ($3$ female, average age $24.7 \pm 5.1$) from the Virginia Tech community. Participants received monetary compensation for their time and gave written consent prior to the start of the experiment under Virginia Tech IRB $\# 23$-$1237$.

The participants provided demonstrations with each method for both tasks. We counterbalanced the order in which the methods were presented to the participants using a Latin Square design (e.g., three participants started with L2D2, three started with Teleop, etc.). Before providing demonstrations with each method, the participants were given $5$ minutes to practice using the joystick or the sketching interface to teach the robot.
Once the participants were familiar with the interface, they provided a total of 5 demonstrations per task. For Teleop, the participants gave 5 teleoperated demonstrations.
For RT-Traj and S2S, they made 5 and 10 drawings, respectively,
while with L2D2, they provided 4 drawings and 1 physical demonstration. 
After using each method, participants answered a survey to report their subjective experience of demonstrating both tasks with that interface.

We combine the demonstrations provided by all users to create the training datasets for each method.

\p{Dependent Variables}
Similar to Section \ref{ss:e2}, we measure the success rate across $10$ task configurations with varying object positions to evaluate the performance of the learned policies. For the \textit{Scooping} task, we compute success by breaking the task into two segments: reaching into the bowl and performing a scooping motion. Each segment accounts for $50\%$ of the task success. Likewise, for the \textit{Pick and Place} task, we create three segments: reaching for the block, grasping and lifting it, and carrying it to the bin, each contributing to a third of the task success. For example, if the robot managed to pick the block but failed to take it to the bin, the policy rollout would be $66.6\%$ successful. 

In addition to the success rate, we also measure the total time taken by each participant to demonstrate both tasks with each method, and their subjective responses to a 7-point Likert Scale survey. The survey questions were arranged into six multi-item scales: how \textit{easy} it was to provide the demonstrations, how \textit{intuitive} the demonstration interface was, and how much \textit{effort} was required to teach the robot. At the end of the study, participants were asked two forced-choice questions: whether they preferred teaching the robot through drawings or teleoperation, and which of the three sketching approaches they preferred to use.

\p{Hypotheses}
We had the following hypotheses:\\
\textbf{H1.} \textit{L2D2 will perform similarly to Teleop, but will outperform other baselines across both tasks.}\\
\textbf{H2.} \textit{The users will require less time to demonstrate the tasks using L2D2 as compared to the baselines.}\\
\textbf{H3.} \textit{Users will perceive that drawings require less effort as compared to physical demonstrations, and will prefer using L2D2 over other drawing approaches.}

\begin{figure}
    \centering
    \includegraphics[width=1.0\linewidth]{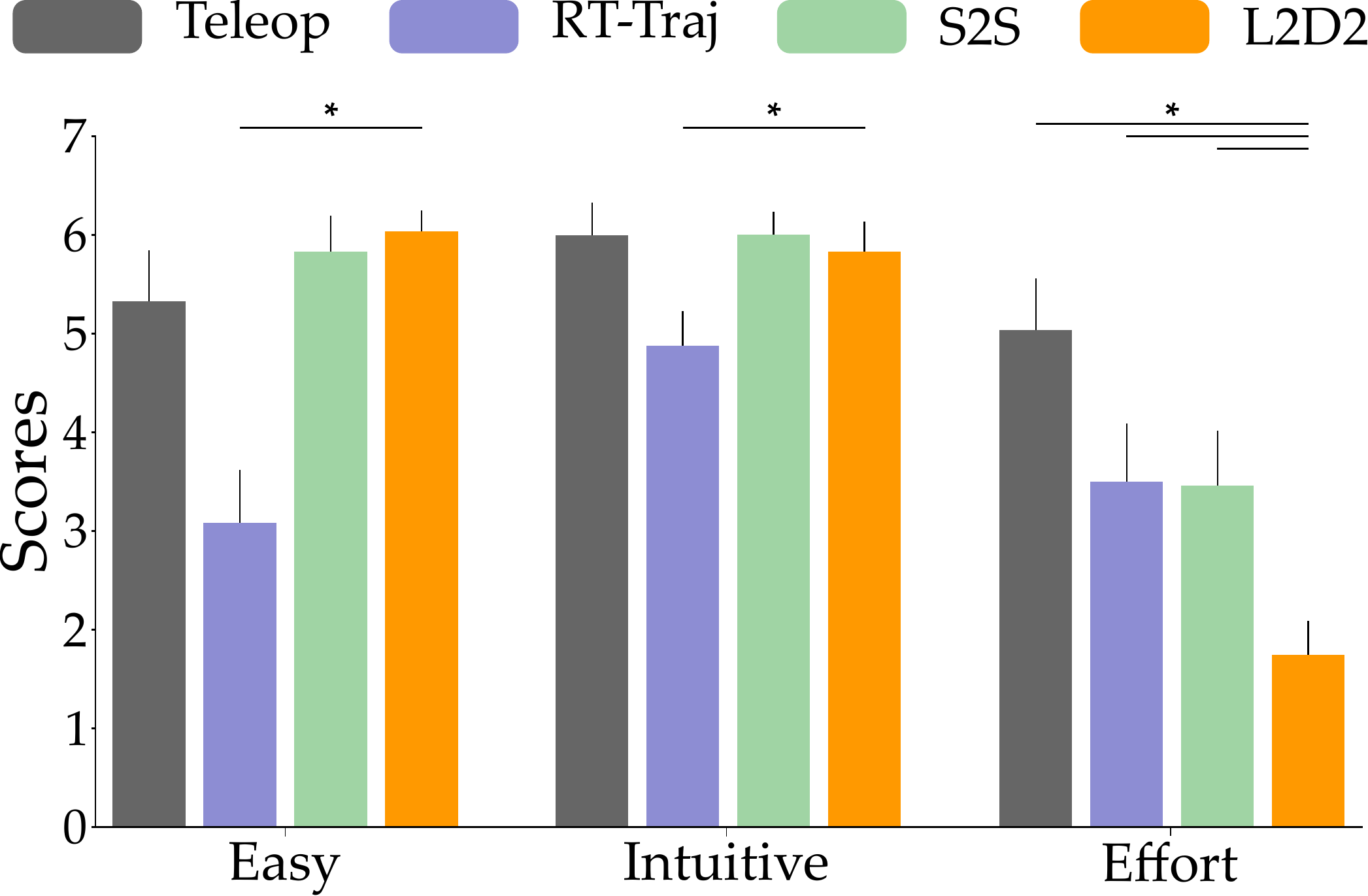}
    \caption{Subjective results from our user study. Higher ratings for \textit{Easy} and \textit{Intuitive} scales represent better subjective experience. On the other hand, for the \textit{Effort} scale, a lower rating indicates that the user had to spend less effort in demonstrating the tasks to the robot. 
    The participants perceived our approach (L2D2) to be as easy and intuitive as teleoperation, and indicated that it requires significantly less effort than all baselines.
    The error bars show the SEM, and $*$ denotes statistical significance $(p<0.05)$.}
    \label{fig:likert}
\end{figure}

\p{Results}
The results for this user study are summarized in Figure \ref{fig:exp2} and Figure \ref{fig:likert}. 

We first evaluate the performance of each method using the data from inexperienced users. A One-way ANOVA revealed that the teaching method had a significant effect on the success rate for the \textit{Scooping} ($F(5,\\ 54) = 9.54, p<0.05$) and \textit{Pick and Place} ($F(5, 54) = 14.22, p<0.05$) tasks.
Post-hoc comparisons for \textit{Scooping} showed that L2D2 outperforms RT-Traj and S2S baselines ($p<0.05$), but performs similarly to Teleop ($p=0.69$). Likewise, for the \textit{Pick and Place} task, L2D2 achieves a similar success rate to Teleop ($p=0.14$) and outperforms all other baselines ($p<0.05$), providing support for our hypothesis \textbf{H1}. This also indicates that the end-users were able to leverage our approach to effectively teach the robot with just 5 minutes of practice.

While L2D2 performs on par with Teleop, its benefits become apparent when we compare the total time taken by users to provide the same amount of data with each approach. 
A repeated measures ANOVA with Greenhouse-Geisser correction revealed that the methods had a significant effect on the overall time spent in demonstrating the task ($F(1.421, 15.63) = 119.99, p<0.05$).  Post hoc analysis also indicated that users spent significantly less time demonstrating the tasks using L2D2 than all baselines ($p<0.05$). This result supports hypothesis \textbf{H2} and demonstrates that L2D2 achieves both learning efficiency and task performance by integrating synthetic drawings with physical feedback.

We next analyze the subjective results from the Likert scale survey summarized in Figure \ref{fig:likert}. After verifying that our scales were reliable (Cronbach's $\alpha > 0.7$), we grouped the responses for each scale. We performed repeated measures ANOVA for each scale with the necessary corrections for violation of sphericity.
The tests indicated that the teaching method had a significant effect on the ease ($F(3, 36) = 11.14$, $p<0.05$) and intuitiveness ($F(2.01,$ $ 24.1$), $p<0.05$) of providing demonstrations. 
Post-hoc comparisons revealed that users fou-nd L2D2 to be as easy and intuitive to use as Teleop, despite not being able to see the robot and objects move in our interface.
Lastly, the test also showed a significant effect of method on the perceived effort ($F(2.2, 26.41) = 13.61$, $p<0.05$), with post hoc comparisons indicating that users put significantly less effort for L2D2 than the baselines ($p<0.05$). We believe that this is because participants had to manually change the positions of task-relevant objects for all baselines, while our approach leveraged vision-language models to automatically vary the object positions in the camera images. 


Overall, $66.6\%$ of users reported that they preferred using drawings to teach the robot as compared to using direct teleoperation, and $91.67\%$ of users stated that they would prefer using L2D2 over other sketching approaches. These subjective results support our hypothesis \textbf{H3}.
Users particularly disliked RT-Traj, stating that \textit{``specifying the height of keypoints is not clear''} in their open-ended responses. Due to this ambiguity, users were unable to select the correct heights, resulting in rollouts where the robot failed to grasp the block or rotate the spoon.
While users did not face the same problem with S2S, the interface did not allow them to specify end-effector rotation for the \textit{Scooping} task, and their drawings for the two camera frames were often misaligned in \textit{Pick and Place}, 
leading to poor reconstruction and task performance.
Instead of obtaining heights or drawings from two viewpoints, our approach used a few teleoperated demonstrations to better reconstruct the robot's trajectory from sketches, which made our interface much more intuitive for end-users and resulted in more accurate learning.


\subsection{Long Horizon Task} \label{ss:e3}

So far, we have demonstrated that our approach works with both expert and novice users. However, our evaluations have only included short manipulation tasks with a single object in the environment. This makes it easy to draw paths that go from the robot's end-effector to the object of interest. But can our approach work for \textit{long-horizon} tasks where the robot must sequentially interact with multiple objects? In this section, we explore whether our sketch-based approach (L2D2) can be leveraged to teach these \textit{long-horizon} tasks. Specifically, we compare the performance of L2D2 to the best-performing baseline from the previous sections (Teleop). 

\p{Task Description}
The experts were tasked with teaching a \textit{Long Horizon} task of setting up a dining table. This task involved sequential interactions with two objects in the environment --- an empty bowl and a can of food --- which can be broken down into two subtasks.
In the first subtask, the robot had to move a bowl to the center of the table, similar to the \textit{Push} task from Section \ref{ss:e1}. Then, in the second subtask, the robot had to pick up the can and place it next to the bowl. The bowl was randomly initialized on the table while the can started in a fixed location.

\begin{figure}
    \centering
    \includegraphics[width=1.0\linewidth]{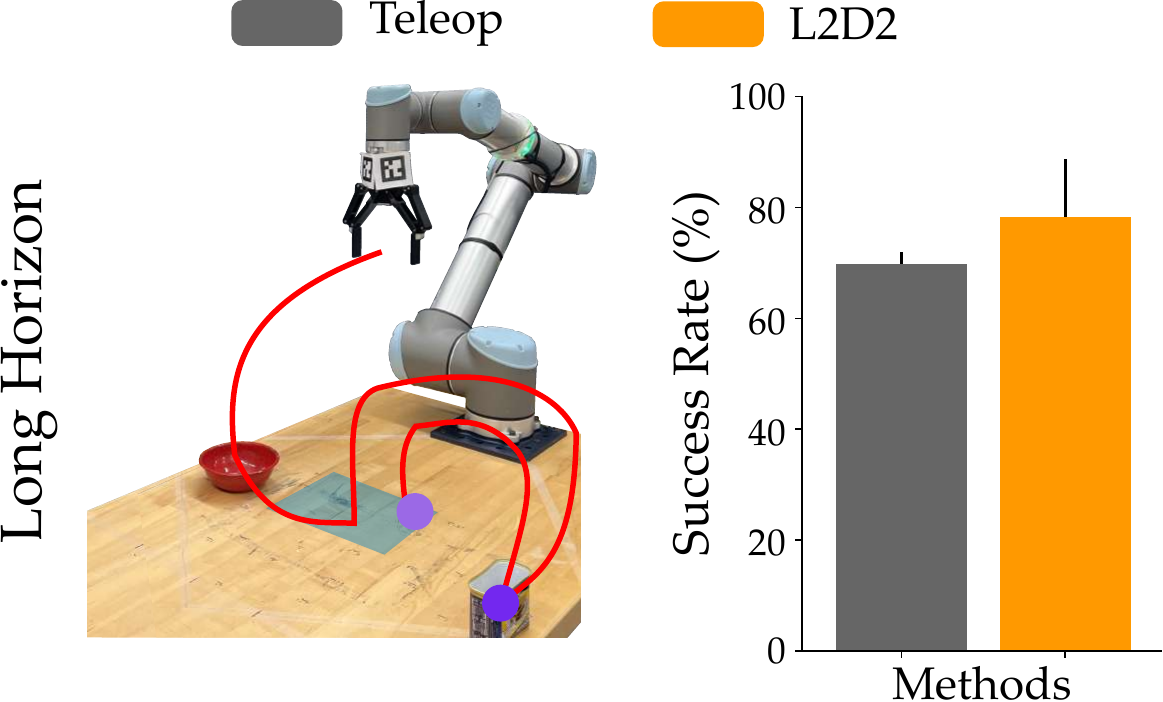}
    \caption{Experimental results for the long-horizon task in Section \ref{ss:e3}. The expert's goal is to teach the robot to push a bowl to the center of the table, followed by picking up a can and placing it next to the bowl. (Left) Environment setup and an example of the drawing provided by the expert. (Right) The success rate of the policies learned from demonstrations collected using Teleop and L2D2. We observe a similar performance for both approaches. Error bars show the SEM.}
    \label{fig:exp3}
\end{figure}

\p{Data Collection}
Similar to Section \ref{ss:e1}, we conduct this experiment with data collected from expert users. For this task, the experts provided a total of $125$ demonstrations to the robot. For Teleop, the experts provided a total of $125$ physical demonstrations by controlling the robot's end-effector velocity using a joystick. On the other hand, for L2D2, the experts provided $100$ drawings and $25$ physical demonstrations to the robot.

\p{Dependent Variables}
Consistent with the previous sections, we evaluate the performance of the robot in terms of success rate across $10$ independent task settings. We compute the success rate by breaking each subtask into smaller segments. We divide the first subtask of moving the bowl into two segments: reaching for the bowl and pushing it to the center of the table, where each segment accounts for $50\%$ of the subtask success. The next subtask of shifting the can is broken into three segments: reaching for the can, grasping and lifting it, and placing it next to the bowl, each representing $33.3\%$ of the subtask success. For example, if the robot completes the first subtask and reaches the can, but fails to close its gripper to grasp it, the rollout is considered to be $66.6\%$ successful.

\p{Results}
The results for this experiment are summarized in Figure \ref{fig:exp3}. An independent samples t-test did not reveal a significant difference in the success rates of policies trained with L2D2 and Teleop ($t(18) = 11.37$, $p=0.441$). This indicates that our sketching-based approach is also effective in \textit{long-horizon} tasks, achieving a performance similar to training with an equivalent dataset of accurate real-world demonstrations.

\section{Conclusion} \label{sec:conclusion}

In this manuscript we proposed L2D2, a drawing-based interface for imitation learning.
Throughout our work we recognize that teaching robot arms by sketching the desired trajectory brings advantages and disadvantages.
The key advantage is the ability to rapidly collect low-effort human demonstrations (i.e., drawings).
The disadvantages stem from the fundamental gap between drawing the task in a static, 2D image and actually performing that same task in our dynamic, 3D world.

To maximize the \textit{potential} of learning from drawings, we leveraged vision and language models to segment the initial image and generate a corpus of synthetic environments.
Human teachers could rapidly draw on multiple of these images to convey diverse examples of the intended task (e.g., showing the robot arm how to grasp a block at various positions on the table).
This ultimately resulted in a sketching interface that users seamlessly interacted with to draw and annotate their desired task.
Participants perceived our method for teaching by drawing to be easier, more intuitive, and less taxing than state-of-the-art baselines.

To minimize the \textit{weaknesses} that are inherent to drawn demonstrations, we developed a two-part solution.
First, we derived an optimization approach based on Principal Component Analysis to place the robot's camera.
By drawing on the images obtained from this camera location, L2D2 mitigated information loss between the user's 2D sketch and the robot's 3D workspace.
Next, to address the gap between static images and dynamic interactions, we grounded the human's drawings by collecting a small set of physical demonstrations.
Using these demonstrations L2D2 refined its understanding of what the drawings meant, and reached a policy that captured how objects should transition.
Our experiments across short- and long-horizon tasks indicated that L2D2 learned a policy that was as effective as policies trained with teleoperated data --- but required less than $75\%$ of the time that users spent in collecting physical demonstrations.

\p{Limitations}
Our results suggest that the benefits of L2D2 increase as the number of demonstrations required to teach the robot scales up: with novice users, providing $100$ demonstrations takes $20$ minutes less with L2D2 than with traditional teleoperation.
We therefore see L2D2 as an important step towards intuitively teaching robot arms to perform real-world tasks.
However, one limitation of L2D2 is that users need to translate the behavior they want the robot to perform into a drawing --- and this can be challenging within long-horizon settings. 
For instance, if the robot needs to manipulate multiple objects, the human illustrator may be unsure what object the robot is currently holding when completing their drawing.
Similarly, if two objects are dynamically coupled in the task (e.g., the robot is pushing a plate with silverware on top of that plate), the drawing may not fully capture the interactions between these objects.
Our current solutions --- including grounding the drawings with real-world data --- help to address this challenge.
But future work should ensure that the visualization of the task captures object manipulation, and that users understand how the gaps in their drawings can be corrected through physical interventions.

\section{Declarations}

\p{Funding} This research was supported in part by the
National Science Foundation (NSF), Grants $\#2205241$ and $\#2337884$.

\p{Conflict of Interest} The authors declare that they have no conflicts of interest.

\p{Ethical Statement} All physical experiments that relied on interactions with humans were conducted under university guidelines and followed the protocol of Virginia Tech IRB $\#23$-$1237$.

\p{Author Contribution} S.M. led the overall algorithm development and wrote the first manuscript draft. H.N. analyzed the algorithm and edited the manuscript. H.S. implemented the baselines and created the drawing interface. S.M., H.N., and H.S. all contributed to the experiments and user study. D.L. supervised the project.

\bibliographystyle{spmpsci}
\bibliography{citations}

\end{document}